%% file: main.tex
\documentclass[11pt]{article}

\usepackage[preprint]{acl}

\usepackage{times}
\usepackage{latexsym}

\usepackage[T1]{fontenc}

\usepackage[utf8]{inputenc}

\usepackage{microtype}

\usepackage{inconsolata}

\usepackage{graphicx}
\usepackage{hyperref}
\usepackage{url}
\usepackage{newfloat}
\usepackage{listings}
\usepackage{amsmath}
\usepackage{booktabs}
\usepackage{multirow}
\usepackage{caption} 
\usepackage{upgreek}
\usepackage{amssymb}
\usepackage{mathtools}
\usepackage{amsthm}
\usepackage{tabularx}
\usepackage[table]{xcolor}
\usepackage{wrapfig}
\newtheorem{theorem}{Theorem}
\usepackage{fontawesome}
\usepackage{siunitx}
\usepackage{enumitem}
\ifodd 0

\newcommand{\com}[1]{\textbf{\color{red}(COMMENT: #1)}}

\newcommand{\clar}[1]{\textbf{\color{green}(NEED CLARIFICATION: #1)}}

\else

\newcommand{\com}[1]{}

\newcommand{\clar}[1]{}
\newcommand{\ty}[1]{}
\newcommand{\tx}[1]{}
\newcommand{\sr}[1]{}
\fi

\definecolor{darkblue}{rgb}{0, 0, 0.5}
\hypersetup{colorlinks=true, citecolor=darkblue, linkcolor=darkblue, urlcolor=darkblue}

\title{Restoring General and Improving Domain-Specific Safety in Fine-Tuned LLMs Without Retraining}

\author{
 \textbf{Vishnu Kabir Chhabra\textsuperscript{1}},
 \textbf{Ding Zhu\textsuperscript{1}},
 \textbf{Zhihui Zhu\textsuperscript{1}},
 \textbf{Xueru Zhang\textsuperscript{1}},
\\
 \textbf{Mohammad Mahdi Khalili\textsuperscript{1}}
\\
\\
\\
 \textsuperscript{1}The Ohio State University, Columbus, OH
\\
 \small{
   \textbf{Correspondence:} \href{mailto:chhabra.67@osu.edu}{chhabra.67@osu.edu}
 }
}

\begin{document}
\maketitle
\begin{abstract}
Safety-aligned language models are known to suffer degradation in refusal behavior following domain-specific fine-tuning, even when the fine-tuning data is entirely benign. The predominant mitigation strategy, preference optimization on the fine-tuned model, is computationally expensive and requires domain-specific preference data that is often unavailable in practice. We propose \textbf{SPECTRA} (\textbf{Spec}tral \textbf{R}efusal \textbf{A}lignment), a training-free, weight-level method that restores refusal behavior by transferring the base model's refusal direction into the fine-tuned model's weights via an activation-aware, low-rank update, requiring no gradient-based optimization or domain-specific safety data. The low-rank design restricts the intervention to the subspace most causally active on harmful inputs, leaving benign behavior largely undisturbed by construction. Empirically, SPECTRA substantially improves refusal behavior across a comprehensive safety evaluation, while preserving general coherence, domain-specific knowledge, and producing negligible change in false refusal rates.
\end{abstract}
\input{sections/intro/intro}

\input{sections/intro/background}

\input{sections/intro/methodology}

\input{sections/intro/exp}

\input{sections/intro/direction}

\input{sections/intro/results}

\input{sections/appendix/frcomp}

\input{sections/appendix/multiturn}

\input{sections/intro/limit}
\input{sections/intro/acknowledgement}

\bibliography{custom}

\appendix

\input{sections/intro/discussion}

\input{sections/intro/related}
\input{sections/appendix/alpha}

\input{sections/appendix/rank}
\input{sections/appendix/rd}
\input{sections/appendix/GCG}
\input{sections/appendix/zero}

\input{sections/appendix/direction}
\input{sections/appendix/ref_data}

\end{document}

%% file: sections/intro/intro.tex
\section{Introduction}
Modern transformer-based large language models (LLMs) \citep{touvron2023llama, llama3modelcard, team2024gemma, team2023gemini} are typically safety-aligned before deployment to ensure safe and trustworthy behavior \citep{rafailov2023direct, ethayarajh2024ktomodelalignmentprospect, bai2022training}. Recent work has shown that preference-based alignment can lead to meaningful improvements across multiple dimensions of model safety \citep{alami2024alignmentpreferenceoptimizationneed, zhang2025bifactorialpreferenceoptimizationbalancing}. Despite this progress, deploying LLMs in specialized settings often requires domain-specific fine-tuning \citep{cheng2024adaptinglargelanguagemodels, anisuzzaman2025fine}, which can inadvertently degrade the very safety properties instilled during alignment, even when the fine-tuning data is entirely benign \citep{qi2023fine, fraser2025fine}. This tension between domain adaptation and safety preservation presents a fundamental challenge for the practical deployment of aligned models.

The vulnerability of aligned models to fine-tuning is well documented. Even a small amount of adversarial data in the fine-tuning set can cause a dramatic drop in safety, while fine-tuning on benign data alone has been shown to erode refusal behavior \citep{qi2023fine, fraser2025fine}. Existing approaches to mitigating this degradation include additional preference optimization \citep{han2024medsafetybenchevaluatingimprovingmedical}, curating domain-specific safety data for further tuning \citep{jan2024multitaskmayhemunveilingmitigating}, modifying the fine-tuning objective \citep{qi2024safetyalignmentjusttokens}, embedding safety backdoors via prefix prompts \citep{NEURIPS2024_094324f3}, and preventing alignment drift through optimization-state separation \citep{huang2024lazy}. Although effective, these methods typically require additional gradient-based computation, modified training procedures, or domain-specific calibration data, which limits their practicality in resource-constrained or data-scarce settings.

A complementary line of research in mechanistic interpretability has revealed that safety-relevant behaviors in language models are encoded as linear directions in the residual stream. \citet{arditi2024refusallanguagemodelsmediated} identified a refusal direction and showed that ablating it from the model's weights constitutes an effective white-box jailbreak, demonstrating that refusal behavior is concretely encoded in weight-space structure. Subsequent work has discovered related steering vectors via sparse autoencoders \citep{o2024steering, cunningham2023sparse} and applied them at inference time to elicit or suppress specific behaviors \citep{obrien2025steeringlanguagemodelrefusal, lee2025programmingrefusalconditionalactivation, panickssery2023steering}. However, inference-time activation steering has fundamental limitations: it tends to increase false refusal rates \citep{obrien2025steeringlanguagemodelrefusal}, requires careful per-model hyperparameter tuning, and introduces runtime overhead that complicates deployment.

These two lines of research together motivate our approach. The problem of fine-tuning-induced safety degradation calls for a lightweight, \textbf{training-free} remedy, while mechanistic interpretability offers tools for identifying and manipulating the weight-space structures that mediate refusal behavior. Rather than retraining the model or steering activations at inference time, we propose to directly inject the base model's refusal direction into the weights of the fine-tuned model. To avoid indiscriminate perturbations to model parameters, which could degrade unrelated capabilities, the injection is restricted to an \textbf{activation-aware}, \textbf{low-rank} subspace of the weight matrix that is most causally active on harmful inputs. This ensures the intervention strengthens refusal behavior on harmful content while remaining largely dormant on benign inputs, without requiring any runtime modification.
We refer to the resulting method as \textsc{SPECTRA} (\textbf{Spec}tral \textbf{R}efusal \textbf{A}lignment). The key components and findings of our work are as follows:

\begin{enumerate}[leftmargin=*,topsep=0.1cm,itemsep=0cm]
\item We propose \textsc{SPECTRA}, a training-free, closed-form weight-space intervention for restoring safety after domain-specific fine-tuning. The method injects a refusal direction from the safety-aligned base model into the weights of the fine-tuned model, requiring no additional gradient updates, domain-specific safety annotations, or inference-time steering.
\item We develop an activation-aware low-rank update that restricts the intervention to the subspace of the weight matrix most active on harmful inputs. This design strengthens refusal behavior where it is needed while minimizing changes to model behavior on benign inputs.
\item We empirically show that \textsc{SPECTRA} substantially improves safety across multiple fine-tuned models and evaluation benchmarks while preserving linguistic quality and domain-specific capabilities. Notably, these gains are not driven by indiscriminate over-refusal, as false refusal rates remain statistically negligible.
\item We provide evidence that refusal directions transfer across fine-tuned variants of a safety-aligned base model, suggesting that fine-tuning preserves a shared geometric structure for refusal and that safety can be restored through targeted weight-space manipulation rather than full retraining.
\end{enumerate}

%% file: sections/intro/background.tex
\section{Problem Formulation}\label{sec:prob}
\subsection{Background}
\label{background}

\paragraph{Transformer notation and residual-stream activations.}
Consider a decoder-only Transformer \citep{radford2019language,vaswani2017attention} that maps token a sequence of token indices $\mathbf{t}=(t_1,\ldots,t_n)\in\mathcal{V}^n$ to token-level output distributions $\mathbf{y}\in\mathbb{R}^{n\times|\mathcal{V}|}$.
Let $\mathbf{x}_i^{(l)}\in\mathbb{R}^{d_{\text{model}}}$ denote the residual-stream activation for token position $i$ at the input of layer $l$. The embedding layer initializes $\mathbf{x}_i^{(1)}=\mathtt{Embed}(t_i)$. For each layer $l\in[L]$, the model applies causal self-attention followed by an MLP, each with a residual connection:
\begin{align}
\mathbf{h}_i^{(l)} &= \mathbf{x}_i^{(l)} + \mathtt{Attn}^{(l)}(\mathbf{x}_{1:i}^{(l)}), ~~~~~\ \\
\mathbf{x}_i^{(l+1)} &= \mathbf{h}_i^{(l)} + \mathtt{MLP}^{(l)}(\mathbf{h}_i^{(l)}).
\end{align}
The final residual-stream activations are projected back to vocabulary space via $\mathtt{logits}_i=\mathtt{Unembed}(\mathbf{x}_i^{(L+1)})$, producing $\mathbf{y}_i=\mathtt{softmax}(\mathtt{logits}_i)$ \citep{arditi2024refusallanguagemodelsmediated}.

\paragraph{From activations to a refusal direction.} A growing body of interpretability work has shown that high-level behavioral properties of language models, including safety-relevant ones such as refusal, are often linearly encoded in the residual stream \citep{marks2023geometry, belrose2023diffinmeans, arditi2024refusallanguagemodelsmediated}. Motivated by this, we seek a single direction in activation space that is causally predictive of refusal behavior. Following the \emph{difference-in-means} methodology \citep{belrose2023diffinmeans, marks2023geometry, arditi2024refusallanguagemodelsmediated}, we compute class-conditional mean activations over a labeled training set of harmful and harmless prompts. For layer $l$ and token position $i$:
\begin{align*}
 \boldsymbol{\upmu}_i^{(l)} &= \mathbb{E}_{\mathbf{t} \sim \mathcal{D}_{\text{harmful}}^{\text{(train)}}} \left[ \mathbf{x}_i^{(l)}(\mathbf{t}) \right],~~~~   \\
\boldsymbol{\upnu}_i^{(l)} &= \mathbb{E}_{\mathbf{t} \sim \mathcal{D}_{\text{harmless}}^{\text{(train)}}} \left[ \mathbf{x}_i^{(l)}(\mathbf{t}) \right].
\end{align*} The candidate refusal direction at 
 $(l,i)$ is defined as the centroid difference 
$\mathbf{r}_i^{(l)}=\boldsymbol{\upmu}_i^{(l)}-\boldsymbol{\upnu}_i^{(l)}$.

\paragraph{Selection the best refusal direction from the cadidates}
The construction above yields a family of candidate directions indexed by layer $l\in[L]$ and token position $i\in I$. To select a single, causally effective refusal direction,  we evaluate each candidate through intervention experiments on a held-out validation set. A direction is considered effective if (i) ablating it from activations induced by harmful prompts in $\mathcal{D}_{\text{harmful}}^{\text{(val)}}$ should suppress refusal, and (ii) injecting it into activations induced by harmless prompts in $\mathcal{D}_{\text{harmless}}^{\text{(val)}}$ should elicit refusal. We select the candidate $(i^*,l^*)$ that maximizes this combined causal effect, and denote the resulting direction as  $\mathbf{r}\coloneqq \mathbf{r}_{i^*}^{(l^*)}$, with normalized version $\hat{\mathbf{r}}$.

%% file: sections/intro/methodology.tex
\subsection{Research Objective}

Safety-aligned language models exhibit strong refusal behavior when prompted with harmful requests. However, this behavior is known to degrade after domain-specific fine-tuning \citep{qi2023fine, fraser2025fine}: when a safety-aligned base model $M_1$ is fine-tuned on domain-specific data to obtain $M_2$, the resulting model often exhibits a substantially lower refusal rate on harmful prompts.  This motivates the central question of our work: 
\begin{center}
    \emph{Can we restore or strengthen refusal behavior in a fine-tuned model without any retraining?}
\end{center}

\citet{arditi2024refusallanguagemodelsmediated} showed that ablating the refusal direction directly from a model's output projection weights, rather than from activations at inference time, produces a persistent white-box jailbreak that reliably bypasses safety guardrails. This suggests that refusal behavior is not merely an emergent property of activations, but is concretely encoded in specific weight-space structures. We build on this insight in the opposite direction: rather than removing weight-space structure to weaken safety, we propose to \emph{inject} it to restore safety. Specifically, given a fine-tuned model $M_2$ with degraded refusal behavior, our goal is to transfer the refusal direction of the base model $M_1$ directly into the weight matrices of $M_2$, achieving a \textbf{training-free}, \textbf{weight-level} safety repair that requires no inference-time activation manipulation.

Formally, let $\mathbf{r}_1$ and $\mathbf{r}_2$ be the refusal directions of $M_1$ and $M_2$, respectively. We seek a targeted, gradient-free modification to a subset of $M_2$’s parameters such that refusal behavior improves on harmful prompts from $\mathcal{D}_{\text{harmful}}^{\text{(val)}}$, while performance on benign prompts from $\mathcal{D}_{\text{harmless}}^{\text{(val)}}$ is preserved.

\section{Method: Spectral Refusal Alignment}\label{sec:method}

A naïve implementation of this idea would inject $\mathbf{r}_2$ into the full weight matrices of $M_2$. However, such a full-matrix intervention is problematic.
Recent work has identified the existence of \emph{superweights} in large language models: a small number of individual parameters with disproportionate influence over model behavior \citep{qin2026achillesheelllmsaltering}. Indiscriminate perturbation of weight matrices risks disrupting these superweights, causing catastrophic degradation in capabilities unrelated to safety. This motivates our approach: rather than modifying the full weight matrix, we restrict our intervention to the low-dimensional subspace most causally active on harmful inputs, thereby preserving benign behavior by design. We refer to the resulting method as \textbf{Spec}tral \textbf{R}efusal \textbf{A}lignment (\textbf{SPECTRA}).

Specifically, let $W\in \mathbb{R}^{d_{\text{out}} \times d_{\text{in}}}$ denote a weight matrix of interest in $M_2$ (either an attention output projection or an MLP projection) and let $X_{\text{in}} \in \mathbb{R}^{d_{\text{in}} \times n}$ denote the matrix of pre-projection activations collected by forwarding a set of harmful prompts (e.g., \textsc{Align} dataset \citep{wei2024assessingbrittlenesssafetyalignment} in our experiments) through $M_2$, where each column corresponds to a response token activation. Our method proceeds in two steps.

\paragraph{Step 1: Activation-Aware Low-Rank Decomposition.}
To identify the subspace of $W$ most active on harmful inputs, we follow \citep{wei2024assessingbrittlenesssafetyalignment, hsu2022language, yuan2023asvd} and seek a rank-$\rho$ approximation $\tilde{W}$ that minimizes the distortion to the output specifically on harmful-prompt activations:
$$
\tilde{W} = \arg \min_{\text{rank}\,\tilde{W} \leq \rho} \| WX_{\text{in}} - \tilde{W}X_{\text{in}} \|_F^2.
$$
This is solved by performing SVD on the output matrix $WX_{\text{in}} \in \mathbb{R}^{d_{\text{out}} \times n}$:
$$
U S V^\top \approx WX_{\text{in}},
$$
where $U \in \mathbb{R}^{d_{\text{out}} \times \rho}$ is the orthogonal matrix corresponding to the top $\rho$ left singular vectors. The minimizer in closed form is:
$$
\tilde{W} = \Pi W, ~\text{where}~\Pi = U U^\top
$$
is the orthogonal projector onto the dominant output subspace induced by harmful activations. Intuitively, $\tilde{W}$
captures the component of $W$ that is most "active" when the model processes harmful content, which is precisely the subspace we wish to steer.

\paragraph{Step 2: Spectral Refusal Injection.} Having identified the safety-relevant subspace via $\tilde{W}$, we inject the refusal direction $\mathbf{r}_1$ of base model $M_1$ by modifying
$W$ as follows:
\begin{align}\label{eq:Wsteering}
    W' = W + \alpha \mathbf{r}_1\mathbf{r}_1^\top\tilde{W}
\end{align}
where $\alpha > 0$ is a scalar steering hyperparameter. The update $\alpha \mathbf{r}_1\mathbf{r}_1^\top\tilde{W}$ adds a rank-one perturbation in the direction of $\mathbf{r}_1$ to the weight matrix, but critically, this perturbation is \emph{gated} by $\tilde{W}$: it only modifies the output along $\mathbf{r}_1$ for inputs that activate the harmful subspace $\Pi$, rather than uniformly modifying for all inputs.

\paragraph{Equivalence to Conditional Activation Steering.} The following theorem shows that our weight-level update is equivalent to a \emph{data-adaptive} form of activation steering, whose strength automatically scales with the extent to which the input representation lies in the harmful-activation subspace.
\begin{theorem}[SPECTRA as conditional activation steering]
\label{thm:equivalence}
    Let $x \in \mathbb{R}^{d_{\text{in}}}$ be an input activation, and let
 $h = Wx \in  \mathbb{R}^{d_{\text{out}}}$ denote the corresponding pre-update output. Under the weight update  \eqref{eq:Wsteering}, the modified output satisfies:
 $$h' = h+\beta \cdot \mathbf{r}_1, $$ where $\beta = \alpha\cdot \mathbf{r}_1^\top \Pi h$ is a scalar that depends on the alignment of $h$ with the harmful subspace $\Pi$. 
\end{theorem}

\begin{proof}
Substituting the definition of $W'$ and $\tilde{W}=\Pi W$:
\begin{align}
    h' &= W'\cdot x = Wx+\alpha \mathbf{r}_1\mathbf{r}_1^\top \Pi Wx \notag\\
       &= h+(\alpha\cdot \mathbf{r}_1^\top \Pi h)\mathbf{r}_1 = h+\beta\cdot \mathbf{r}_1,
\end{align}
where $\beta = \alpha\cdot \mathbf{r}_1^\top \Pi h$ is a scalar.
\end{proof}
Theorem~\ref{thm:equivalence} reveals the key property that makes SPECTRA both effective and capability-preserving. When $x$ is induced by a
harmful prompt, $h=Wx$ expected to lie substantially within the span of $U$, since $U$ is constructed from harmful-prompt activations. This yields a large $\beta$ and therefore strong steering toward the refusal direction. Conversely, when $x$ is from a
benign prompt, $h$ is less likely to align with $u$, causing $\beta \approx 0$
 and leaving the output effectively unchanged. In this sense, SPECTRA is \textit{geometrically self-gating}: the strength of the refusal intervention scales automatically with how harmful the input representation is, without requiring any explicit runtime classifier or threshold.

%% file: sections/intro/exp.tex
\section{Experimental Setup}\label{sec:exp_setup}

\paragraph{Models and domain fine-tuned variants.} 

We study safety-aligned, widely used LLMs: \textsc{Llama-2-Chat-7B} \citep{touvron2023llama-2} and \textsc{Gemma2-9B-IT} \citep{team2024gemma}, together with publicly available domain fine-tuned variants. For \textsc{Llama-2-Chat-7B}, we include medical \citep{rohanian2024exploring}, law \citep{cheng2024adaptinglargelanguagemodels}, and finance \citep{cheng2024adaptinglargelanguagemodels} models. For \textsc{Gemma2-9B-IT}, we include medical \citep{OpenMeditron2024} and finance \citep{WiroAI} variants.

\paragraph{Coherence evaluation.}
We evaluate model coherence using zero-shot accuracy on both general-purpose and domain-specific benchmarks. 
\begin{itemize}[leftmargin=*,topsep=0.1cm,itemsep=0.cm]
    \item \textbf{General coherence}: We use five tasks from EleutherAI’s LM Harness \citep{eval-harness}: HellaSwag \citep{zellers2019hellaswag}, BoolQ \citep{clark2019boolq}, RTE \citep{wang2018glue}, ARC Challenge \citep{clark2018think}, Winogrande \citep{sakaguchi2021winogrande}. 
    \item \textbf{Domain coherence}: We use standard benchmarks in each domain: MedMCQA \citep{pal2022medmcqa} and PubMedQA \citep{jin2019pubmedqa} for \textit{medical}, FinanceBench \citep{islam2023financebench} for \textit{finance}, and LegalBench \citep{guha2023legalbench} for \textit{law}; for LegalBench we report results on the Abercrombie and Hearsay tasks.
    \end{itemize}

\paragraph{Safety evaluation and threat models.}
We evaluate safety using attack success rate (ASR)\footnote{We sometimes refer to ASR as the attack score.} under harmful instructions, reporting both general and domain-specific safety. 
\begin{itemize}[leftmargin=*,topsep=0.1cm,itemsep=0.cm]
    \item  \textbf{General safety:} We prompt each model with \textsc{AdvBench-eval} (the first 100 prompts from \textsc{AdvBench}) and compute ASR as the fraction of prompts for which the model’s response does not exhibit refusal patterns, following \citet{zou2023universal}. Following \citet{wei2024assessingbrittlenesssafetyalignment}, we evaluate two threat models: standard decoding (ASR$_\mathrm{Vanilla}$) and adversarial decoding (ASR$_\mathrm{Adv\text{-}Decoding}$) \citep{huang2023catastrophic}, where the attacker manipulates the decoding process. For ASR$_\mathrm{Adv\text{-}Decoding}$, we additionally report results with and without the chat template wrapper, denoted ASR$_\mathrm{Adv\text{-}Decoding}^{I}$ and ASR$_\mathrm{Adv\text{-}Decoding}^{\times}$, respectively.\footnote{\textsc{Gemma2-9B-IT} doesn't use \texttt{[INST]} wrapper; we follow the prompt template in \citet{team2024gemma}.}
    
    \item \textbf{Domain safety:} For the \textit{medical} domain, we evaluate ASR on MedSafeEval \citep{han2024medsafetybenchevaluatingimprovingmedical}. For \textit{law} and \textit{finance}, where standardized domain safety benchmarks are limited, we construct \textbf{LawSafeEval} (201 prompts) and \textbf{FinSafeEval} (142 prompts): we first generate 2{,}000 candidate safety-related prompts per domain by prompting Gemini 2.5 Pro \citep{comanici2025gemini} with domain-specific safe prompts as references, and then manually filter and label candidates to select the final prompt sets.
\end{itemize}

\paragraph{Jailbreak robustness (adaptive safety evaluation).}
In addition to ASR on fixed prompt sets, we evaluate robustness to adaptive jailbreak attacks, including GCG \citep{zou2023universal}, GPTFuzz \citep{yu2023gptfuzzer}, and TAP \citep{mehrotra2024tree}.

\paragraph{Refusal labeling and ASR computation.}
 We use the \texttt{HarmBench Llama2-13B Classifier} \citep{mazeika2024harmbench} to label outputs as refusal (refusal-score $=1$) or successful attack (refusal-score $=0$), following prior work \citep{lermen2023lora,liu2023autodan,robey2023smoothllm,shah2023loft,xu2023cognitive,zou2023universal}. For TAP \citep{mehrotra2024tree} and GPTFuzz \citep{yu2023gptfuzzer}, we use GPT-3.5-Turbo \citep{openai2023gpt4} to assign refusal scores.

\paragraph{Datasets for estimating refusal direction and safety-critical ranks.}
In addition to evaluation benchmarks, our method uses auxiliary datasets to estimate (i) a refusal direction in activation space and (ii) safety-critical low-rank structure in fine-tuned weight matrices.
\begin{itemize}[leftmargin=*,topsep=0.1cm,itemsep=0cm]
    \item \textbf{Refusal direction estimation.}
We follow the procedure outlined by \citet{arditi2024refusallanguagemodelsmediated}, see \autoref{sec:ref_data}.
\item \textbf{Safety-critical ranks isolation.}
We use the \textsc{Align} \cite{wei2024assessingbrittlenesssafetyalignment} dataset to isolate safety-critical ranks in fine-tuned weight matrices. 
\end{itemize}

%% file: sections/intro/direction.tex
\section{On the efficacy of refusal direction for inducing safety}\label{sec:simple}

We first ask which refusal direction to inject, and into which layer. Let $\mathbf{r}_1$ be the refusal direction extracted at layer $l_1$ of the base model $M_1$, and $\mathbf{r}_2$ the one extracted at layer $l_2$ of the fine-tuned model $M_2$; fine-tuning can shift the layer and token position at which the direction is found (Appendix~H.1). The two natural choices are to steer layer $l_1$ with $\mathbf{r}_1$, or layer $l_2$ with $\mathbf{r}_2$. We additionally test steering layer $l_2$ with $\mathbf{r}_1$, since the layer that carries refusal may move during fine-tuning while the base model's direction can still apply. We also include $\mathbf{r}_2-\mathbf{r}_1$ as a control that combines both directions.
 
To isolate this choice, we inject the direction into the full attention-output and MLP projection weights of the target layer, without any low-rank approximation:
\begin{equation}
W' = W + \alpha\, \mathbf{r}\mathbf{r}^\top W,
\label{eq:ablation}
\end{equation}
a special case of SPECTRA (Eq.~\eqref{eq:Wsteering}) with $\tilde{W}=W$. We evaluate every direction--layer pair on Llama-2-7b-chat fine-tuned on medical data, using our standard safety metrics and GCG \citep{zou2023universal} (Table~\ref{tab:saf_ben}).
 
\begin{table}[t]
\setlength{\tabcolsep}{3pt}
\centering
\resizebox{\columnwidth}{!}{
\begin{tabular}{llcccc}
\toprule
Direction & Layer & ASR$_\textrm{Adv-Dec.}^{I}$ & ASR$_\textrm{Vanilla}$ & ASR$_\textrm{Adv-Dec.}^{\times}$ & GCG \\
\midrule
\multicolumn{2}{l}{No steering (base FT)} & 1.4 & 0.0 & 18.4 & 31 \\
\midrule
$\mathbf{r}_1$ & $l_2$ (default) & 2.8 & 0 & \textbf{10.8} & \textbf{4} \\
$\mathbf{r}_1$ & $l_1$ & 3.2 & 1.0 & 11.6 & 20 \\
$\mathbf{r}_2$ & $l_2$ & 2.0 & 0.0 & 16.4 & 25 \\
$\mathbf{r}_2$ & $l_1$ & 1.8 & 1.0 & 18.6 & 27 \\
$\mathbf{r}_2-\mathbf{r}_1$ & $l_2$ & 8.0 & 3.0 & 13.6 & \textbf{4} \\
$\mathbf{r}_2-\mathbf{r}_1$ & $l_1$ & 4.6 & 1.0 & 14.4 & 18 \\
\bottomrule
\end{tabular}}
\caption{Full-matrix injection (Eq.~\ref{eq:ablation}) of each refusal direction into the attention and MLP projections of a given layer, on medical Llama-2-7b-chat. $\alpha{=}0.015$ (attention) and $0.01$ (MLP). Lower is better; bold marks the best value per column among steered variants.}
\label{tab:saf_ben}
\end{table}

Injecting $\mathbf{r}_1$ at $l_2$ gives the lowest ASR$^{\times}_\textrm{Adv-Dec.}$ and, together with $\mathbf{r}_2-\mathbf{r}_1$ at $l_2$, the lowest GCG ASR (4 vs.\ 31 unsteered), while staying near zero on other settings. Among the two $l_2$ variants with low GCG ASR, only $\mathbf{r}_1$ avoids the rise on ASR$_\textrm{Vanilla}$ and ASR$^{I}_\textrm{Adv-Dec.}$ seen with $\mathbf{r}_2-\mathbf{r}_1$. We therefore use $\mathbf{r}_1$ at layer $l_2$ in all subsequent experiments.

\paragraph{Why does the base model's direction work better?}
At the same layer $l_2$, injecting $\mathbf{r}_1$ outperforms injecting $\mathbf{r}_2$. To see why, we compare $M_1$ with $M_2$ (Llama-2-7b-chat and its medical fine-tune) at layer $l_2{=}13$ on 200 harmful \textsc{harmful\_behaviors} prompts \citep{zou2023universal} and 200 harmless Alpaca prompts \citep{taori2023alpaca}, using three measures. \emph{Separation} is the discriminability $d'$ (difference of class means over the pooled standard deviation) between harmful and harmless activations projected onto a direction at the last prompt token. We compare $\mathbf{r}_1$, $\mathbf{r}_2$, and a held-out difference-in-means (DiM) direction computed on a random half of the prompts and scored on the other half. \emph{Write} is the mean of $\mathbf{r}^{\top}Wx$ over harmful prompts, summed over the attention and MLP output projections $W$ of layer $l_2$ with inputs $x$, evaluated at each direction's extraction position (token $-1$ for $\mathbf{r}_1$, $-5$ for $\mathbf{r}_2$). We also compute $\mathbf{r}^{\top}(W_1{-}W_2)x$ on $M_2$'s inputs to separate the effect of the layer's weights from that of its inputs. \emph{Gate selectivity} is the ratio of $|\mathbf{r}^{\top}\Pi Wx|$ on harmful versus harmless prompts, where SPECTRA's update adds $\alpha\,\mathbf{r}\,(\mathbf{r}^{\top}\Pi Wx)$ to the output and $\Pi$ projects onto the harmful subspace; a ratio of 1 means no selectivity.

\begin{figure}[t]
  \centering
  \includegraphics[width=0.6\columnwidth]{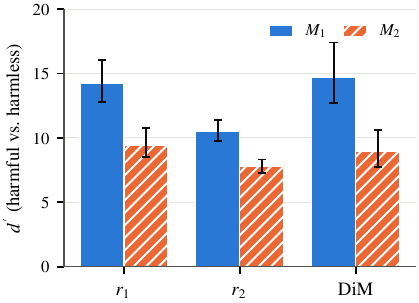}
  \caption{Separation ($d'$) of harmful from harmless prompts at layer $l_2$ and the last prompt token, projected onto $\mathbf{r}_1$, $\mathbf{r}_2$, and DiM, for $M_1$ and $M_2$. }
  \label{fig:signal}
\end{figure}

\begin{figure}[t]
  \centering
  \includegraphics[width=0.6\columnwidth]{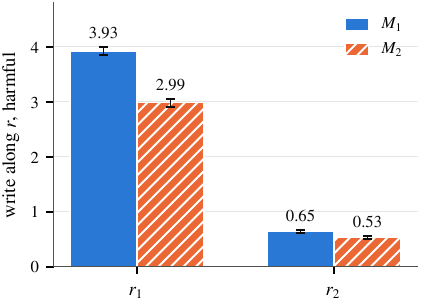}
  \caption{(a)~Write along each direction at its extraction position, over harmful prompts. (b)~Share of the $M_1$--$M_2$ drop in (a) due to the weight difference $W_1{-}W_2$ at $l_2$; the remainder comes from differing inputs.}
  \label{fig:write}
\end{figure}

\begin{figure}[t]
  \centering
  \includegraphics[width=0.6\columnwidth]{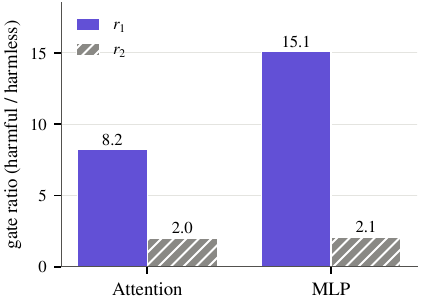}
  \caption{Gate selectivity of the SPECTRA update at layer $l_2$: ratio of the mean $|\mathbf{r}^{\top}\Pi Wx|$ on harmful to harmless prompts, for attention and MLP.}
  \label{fig:gate}
\end{figure}

Fine-tuning lowers separation, but not specifically for $\mathbf{r}_1$: all three directions separate the classes less well in $M_2$ than in $M_1$ (Figure~\ref{fig:signal}). Each direction also loses roughly a quarter of its write in $M_2$ at its own extraction position, and almost none of this loss is due to the weights at $l_2$ (Figure~\ref{fig:write}). The shortfall therefore originates in earlier layers: injection at $l_2$ restores a signal that upstream layers no longer deliver, rather than reverting a weight edit. Since $\mathbf{r}_1$ and $\mathbf{r}_2$ are eroded to a similar degree, erosion does not explain why $\mathbf{r}_1$ is the better direction. The gate does (Figure~\ref{fig:gate}). At the last prompt token, from which the first response token is predicted, the gate is strongly selective for $\mathbf{r}_1$ but only weakly so for $\mathbf{r}_2$, and at token $-5$ it is close to unselective for both; correspondingly, the update along $\mathbf{r}_1$ adds substantially more write at the last token than the update along $\mathbf{r}_2$. The advantage of $\mathbf{r}_1$ thus comes from its alignment with the harmful-specific write pathway at the decisive position, so injecting it acts selectively on harmful inputs, whereas injecting $\mathbf{r}_2$ largely does not.

%% file: sections/intro/results.tex
\begin{table*}[h]
\centering
\resizebox{0.8\linewidth}{!}{
\setlength{\tabcolsep}{4pt}
\renewcommand{\arraystretch}{1.2} 
\begin{tabular}{@{}lllccccccc@{}}
\toprule
\textbf{Model} & \textbf{Domain} & \textbf{Method} 
& \textbf{GCG} & \textbf{AdvDecoding}
& \textbf{Vanilla} & \textbf{GPTFuzz} & \textbf{TAP} & \textbf{In-Domain} \\ 
\midrule
\multirow{6}{*}{Llama2}     
    & \multirow{2}{*}{\textbf{Medical}}   
        & BASE      & $31.0$ & $18.40$ & $0.0$ & $8.8$ & $22.0$ & $28.78$ \\
    &                         & SPECTRA  & $\mathbf{6.0}$ & $\mathbf{9.6}$ & $0.0$ & $\mathbf{0.0}$ & $\mathbf{0.0}$ & $\mathbf{11.56}$ \\
\cmidrule(l){2-9}
    & \multirow{2}{*}{\textbf{Law}}       
        & BASE      & $66$ & $29.6$ & $0.0$ & $15.0$ & $58.0$ & $17.0$ \\
    &                         & SPECTRA  & $\mathbf{43.0}$ & $\mathbf{23.2}$ & $0.0$ & $\mathbf{13.0}$ & $\mathbf{0.0}$ & $\mathbf{6.5}$ \\
\cmidrule(l){2-9}
    & \multirow{2}{*}{\textbf{Finance}}   
        & BASE     & $62.0$ & $29.0$ & $0.0$ & $7.8$ & $28.0$ & $8.45$ \\
    &                         & SPECTRA  & $\mathbf{47.0}$ & $29.0$ & $0.0$ & $7.8$ & $\mathbf{0.0}$ & $\mathbf{7.05}$ \\
\midrule
\multirow{4}{*}{Gemma2} 
    & \multirow{2}{*}{\textbf{Medical}}   
        & BASE      & $36.0$ & $9.4$ & $1.0$ & $68.6$ & $36.0$ & $20.0$ \\
    &                         & SPECTRA  & $\mathbf{0.0}$ & $\mathbf{5.6}$ & $\mathbf{0.0}$ & $\mathbf{7.0}$ & $\mathbf{0.0}$ & $\mathbf{5.44}$ \\
\cmidrule(l){2-9}
    & \multirow{2}{*}{\textbf{Finance}}   
        & BASE      & $66.0$ & $49.2$ & $21.0$ & $64.7$ & $8.0$ & $71.83$ \\
    &                         & SPECTRA  & $\mathbf{1.0}$ & $\mathbf{16.0}$ & $\mathbf{0.0}$ & $\mathbf{0.0}$ & $10.0$ & $\mathbf{7.04}$ \\
\bottomrule
\end{tabular}
}
\caption{ASR (\%) of base fine-tuned models vs.\ SPECTRA across models and domains. Lower is better; best results per domain block are \textbf{bolded}. SPECTRA consistently reduces ASR across attack types, with the most pronounced improvements for Gemma2.} 
\label{tab:rfuc}
\end{table*}

\section{Experimental Results}\label{sec:exp_result}

\begin{table*}[h]
\centering
\resizebox{\linewidth}{!}{
\begin{tabular}{@{}llcc cc c@{}}
\toprule
\textbf{Model} & \textbf{Domain} 
& \multicolumn{2}{c}{\textbf{Medical}} 
& \multicolumn{2}{c}{\textbf{Law}} 
& \textbf{Finance} \\
\cmidrule(lr){3-4} \cmidrule(lr){5-6} \cmidrule(lr){7-7}
& & \textbf{PubMedQA} & \textbf{MedMCQA} & \textbf{Abercrombie} & \textbf{Hearsay} & \textbf{FinanceBench} \\
\midrule
\multirow{3}{*}{Llama2}     
    & Medical   & $77.5$ / \color{gray}{$77.5$}\color{blue}{$(+0.0)$} 
                & \textbf{$41.7$} / \color{gray}{$41.7$}\color{blue}{$(+0.0)$} 
                & --   & --   & --   \\
    & Law       & --   & --   & $29.4$ / \color{gray}{$28.4$}\color{green}{$(+1.0)$} 
                             & $63.0$ / \color{gray}{$65.2$}\color{red}{$(-2.0)$} 
                             & --   \\
    & Finance   & --   & --   & --   & --   & $70.0$ / \color{gray}{$66.7$}\color{green}{$(+3.3)$}  \\
\midrule
\multirow{2}{*}{Gemma2} 
    & Medical   & $95.5$ / \color{gray}{$95.5$}\color{blue}{$(+0.0)$}  
                & $52.5$ / \color{gray}{$52.5$}\color{blue}{$(+0.0)$}  
                & --   & --   & --   \\
    & Finance   & --   & --   & --   & --   & $66.7$ / \color{gray}{$68.0$}\color{red}{$(-1.3)$}   \\
\bottomrule
\end{tabular}
}
\caption{In-domain performance of base fine-tuned models and SPECTRA across models and domains. Medical domains are evaluated on PubMedQA and MedMCQA, Law on Abercrombie and Hearsay from LegalBench, and Finance on FinanceBench.} 
\label{tab:spectra_results}
\end{table*} 

\begin{table*}[h!]
\centering
\resizebox{\linewidth}{!}{
\setlength{\tabcolsep}{2pt}
\renewcommand{\arraystretch}{1.5}
\begin{tabular}{@{}llccccc@{}}
\toprule
\textbf{Model} & \textbf{Domain} & \textbf{RTE} & \textbf{ARC} & \textbf{BoolQ} & \textbf{Winogrande} & \textbf{HellaSwag} \\ 
\midrule
\multirow{3}{*}{Llama2}     
    & \textbf{Medical}   & $68.5$ / \color{gray}{$67.5$}\color{green}{$(+1.0)$} & $45.0$ / \color{gray}{$45.0$}\color{blue}{$(+0.0)$} & $78.0$ / \color{gray}{$78.5$}\color{red}{$(-0.5)$} & $69.0$ / \color{gray}{$68.5$}\color{green}{$(+0.5)$} & $60.0$ / \color{gray}{$58.5$}\color{green}{$(+1.5)$} \\
    & \textbf{Law}       & $69.5$ / \color{gray}{$69.0$}\color{green}{$(+0.5)$} & $44.5$ / \color{gray}{$44.0$}\color{green}{$(+0.5)$} & $78.0$ / \color{gray}{$80.0$}\color{red}{$(-2.0)$}& $65.5$ / \color{gray}{$68.0$}\color{red}{$(-2.5)$} & $57.0$ / \color{gray}{$57.5$}\color{red}{$(-0.5)$} \\
    & \textbf{Finance}   & $70.0$ / \color{gray}{$70.5$}\color{red}{$(-0.5)$} & $41.5$ / \color{gray}{$40.5$}\color{green}{$(+1.0)$} &  $81.0$ / \color{gray}{$79.5$}\color{green}{$(+1.5)$} & $73.5$ / \color{gray}{$70.0$}\color{green}{$(+3.5)$} & $52.0$ / \color{gray}{$52.0$}\color{blue}{$(+0.0)$} \\
\midrule
\multirow{2}{*}{Gemma2} 
    & \textbf{Medical}   & $72.5$ / \color{gray}{$72.0$}\color{green}{$(+0.5)$} & $55.5$ / \color{gray}{$57.5$}\color{red}{$(-2.0)$} & $88.0$ / \color{gray}{$88.0$}\color{blue}{$(+0.0)$} & $72.5$ / \color{gray}{$71.0$}\color{green}{$(+1.5)$} & $52.5$ / \color{gray}{$53.0$}\color{red}{$(-0.5)$} \\
    & \textbf{Finance}   & $77.0$ / \color{gray}{$80.5$}\color{red}{$(-3.5)$} & $53.0$ / \color{gray}{$52.5$}\color{green}{$(+0.5)$} & $86.0$ / \color{gray}{$91.0$}\color{red}{$(-5.0)$} & $70.5$ / \color{gray}{$72.5$}\color{red}{$(-2.0)$} & $53.5$ / \color{gray}{$53.5$}\color{blue}{$(+0.0)$} \\
\bottomrule
\end{tabular}
}
\caption{General coherence benchmark scores across models and domains. For each entry, black denotes the SPECTRA result and {\color{gray}gray} denotes the base fine-tuned model result, with {\color{green}green} indicating improvement and {\color{red}red} indicating degradation relative to the base model.}
\label{tab:reval}
\end{table*}

\subsection{Safety Evaluations}
\paragraph{General Safety.} We evaluate the general safety of models steered via SPECTRA against their base fine-tuned counterparts. Table~\ref{tab:rfuc} reports ASR scores (in \%)\footnote{For \texttt{Llama-2-7b-chat}, we report ASR$_\textrm{Adv-Decoding}^{\times}$ due to the high safety risks of the method.}. It demonstrates that SPECTRA substantially improves refusal behavior across all evaluated models, while remaining lightweight and requiring no retraining. The improvement is most pronounced for both variants of \texttt{Gemma-2-9b-it}, where SPECTRA yields the largest reductions in ASR. We note that each model and its fine-tuned variants require distinct values of $\alpha$ and low-rank projection rank $\rho$; mild hyperparameter tuning was performed to obtain the reported results. We refer the reader to Appendix~\ref{app:alpha} and~\ref{app:rank} for the full parameter choices, and to Appendix~\ref{app:rdir} for the layers on which steering was applied.

\paragraph{Robustness Against Jailbreaks.} Table~\ref{tab:rfuc} shows that SPECTRA substantially increases each model variant's robustness against jailbreak attacks. Importantly, this improvement is \textit{method-agnostic}: SPECTRA safeguards the model against both white-box attacks that require access to model weights, such as GCG \citep{zou2023universal}, and black-box attacks that do not, such as GPTFuzz \citep{yu2023gptfuzzer} and TAP \citep{mehrotra2024tree}.  

\paragraph{In-Domain Safety.} SPECTRA also yields strong improvements in domain-specific safety, notably without requiring any domain-specific data in its procedure. To the best of our knowledge, SPECTRA is the first training-free method to improve the in-domain safety of fine-tuned language models without any backpropagation, making it uniquely practical for deployment settings where retraining is infeasible.

\subsection{Coherence Evaluations}

\paragraph{General Coherence.} We evaluate general coherence via zero-shot performance on HellaSwag, WinoGrande, ARC Challenge, and BoolQ. As shown in Table~\ref{tab:reval}, the impact of SPECTRA on general coherence is statistically negligible, confirming that safety restoration comes at little cost to general capability. In some cases, SPECTRA marginally improves benchmark performance; while this effect is minor, it suggests an interesting direction for future work exploring whether weight-level steering can simultaneously improve both safety and capability. The most notable exception is \texttt{Gemma-2-9b-it} fine-tuned on finance data, which exhibits the largest decrease in general coherence across all tested models; however, this decline is mild and is accompanied by the largest gain in safety performance of any model in our evaluation, representing a favorable overall trade-off.

\paragraph{In-Domain Coherence.} To further assess the effect of SPECTRA on task utility, we evaluate domain-specific knowledge retention for all models modified by SPECTRA, with results reported in Table~\ref{tab:spectra_results}. We find that these models retain their domain-specific knowledge, with no statistically significant decrease in domain performance. Together with the general coherence results above, this confirms that SPECTRA introduces \textbf{no significant degradation in either general or domain-specific capability}, achieving a favorable safety--utility trade-off without any retraining.

\subsection{False Refusal and Comparison to other methods}
A known limitation of refusal steering is its tendency to trigger refusal on unrelated, benign prompts \citep{obrien2025steeringlanguagemodelrefusal,lee2025programmingrefusalconditionalactivation}. To assess whether this limitation applies to SPECTRA, we measure false refusal rates before and after applying SPECTRA across all model variants.

Concretely, we prompt each model with 1000 benign instructions from the \texttt{Alpaca} dataset \citep{taori2023alpaca} and compute the percentage of responses that constitute false refusals. Model responses are evaluated using Gemini 2.5 Pro \citep{comanici2025gemini} as an automated judge, with results reported in Table~\ref{tab:long_format_comparison}. We find that SPECTRA produces negligible change in false refusal rates across nearly all model variants. The one exception is \texttt{Gemma-2-9b-it} fine-tuned on finance data, which exhibits a small increase of 1.3\%, notably the same model that achieves the largest safety gains across all evaluated models, suggesting a mild safety--utility trade-off in this case. Crucially, unlike activation-based steering methods \citep{obrien2025steeringlanguagemodelrefusal}, \textbf{SPECTRA does not meaningfully increase false refusal rates}.

Furthermore, we compare our methodology to prior methods that utilize steering vectors/ model editing to induce safety in post-trained models. We specifically focus on prior literature that aims to induce refusal directions such as Activation Addition (ActAdd) \cite{arditi2023refusal}, ROSI \cite{shairah2025turning}, CAST \cite{lee2025programmingrefusalconditionalactivation}, Steer2Edit \cite{sun2026steer2editactivationsteeringcomponentlevel} and a model editing methodology,SafeMERGE, that is aimed to specifically aid fine-tune models safety \cite{Djuhera_2026} and present the comparison in \autoref{tab:long_format_comparison}. 

\definecolor{ours}{RGB}{225,236,250}
\newcommand{\best}[1]{\mathbf{#1}}

\begin{table}[h]
\footnotesize
\setlength{\tabcolsep}{3pt}
\renewcommand{\arraystretch}{1.05}
\centering
\resizebox{\columnwidth}{!}{
\begin{tabular}{ll ccccc >{\columncolor{ours}}c}
\toprule
\textbf{Model} & \textbf{Domain} & \textbf{ActAdd} & \textbf{ROSI} & \textbf{CAST} & \textbf{Steer2Edit} & \textbf{SafeMERGE} & \textbf{SPECTRA} \\
\midrule
\multirow{2}{*}{Gemma2}
& Finance & $91.2$ & $7.3$ & $55.8$ & $18.8$ & $16.4$ & $\best{1.3}$ \\
& Medical & $88.9$ & $5.7$ & $65.9$ & $17.4$ & $11.2$ & $\best{0.3}$ \\
\midrule
\multirow{3}{*}{Llama2}
& Finance & $93.4$ & $9.5$ & $61.8$ & $22.5$ & $13.7$ & $\best{0.1}$ \\
& Law     & $97.9$ & $6.9$ & $62.4$ & $21.6$ & $19.1$ & $\best{-0.8}$ \\
& Medical & $98.5$ & $8.1$ & $63.8$ & $20.4$ & $14.4$ & $\best{0.2}$ \\
\midrule
\multicolumn{2}{l}{Average} & $94.0$ & $7.5$ & $61.9$ & $20.1$ & $15.0$ & $\best{0.2}$ \\
\bottomrule
\end{tabular}}
\caption{Change in false refusal rate (Post $-$ Pre, in \%) after steering. Lower is better; best per row in bold.}
\label{tab:long_format_comparison}
\end{table}

%% file: sections/appendix/frcomp.tex
%% file: sections/appendix/multiturn.tex
\section{Impact of SPECTRA on Multi-Turn Dialogue}
\label{app:multi_turn}
Prior sections establish that SPECTRA has minimal impact on both general and in-domain task performance. To probe more subtle, fine-grained effects, we benchmark SPECTRA on \textbf{MultiChallenge} \citep{sirdeshmukh2025multichallengerealisticmultiturnconversation}, a multi-turn conversational evaluation designed to assess capabilities such as inference memory, self-coherence, instruction retention, and reliable editing. As shown in Table~\ref{tab:spectra_comparison}, the effect of SPECTRA on multi-turn capabilities varies across models, domains, and evaluation dimensions, with no consistent directional trend. We note that the base models themselves achieve relatively low scores on this benchmark, which limits the strength of conclusions that can be drawn from these results alone. Evaluating SPECTRA on larger, more capable models would be necessary to generalize these findings, and we identify this as a promising direction for future work.

\begin{table}[h]
    \centering
    \renewcommand{\arraystretch}{1.2}
    \setlength{\tabcolsep}{4pt}
\resizebox{\columnwidth}{!}{
    \begin{tabular}{lllcccc}
        \toprule
        \textbf{Domain} & \textbf{Model} & \textbf{Method} & \textbf{Inference} & \textbf{Self} & \textbf{Instruction} & \textbf{Reliable} \\
                        &                &                 & \textbf{Memory}    & \textbf{Coherence} & \textbf{Retention} & \textbf{Editing} \\
        \midrule
        \multirow{4}{*}{\textbf{Finance}}
        & \multirow{2}{*}{Gemma2} & Base    & 2.15 & 9.09  & 29.41 & 0.00 \\
        &                         & SPECTRA & 4.42 & 10.00 & 13.04 & 5.00 \\
        \cmidrule(lr){2-7}
        & \multirow{2}{*}{Llama2} & Base    & 4.42 & 2.04  & 14.49 & 14.63 \\
        &                         & SPECTRA & 5.31 & 10.00 & 5.80  & 9.76  \\
        \midrule
        \multirow{4}{*}{\textbf{Medical}}
        & \multirow{2}{*}{Gemma2} & Base    & 8.85 & 6.00  & 10.14 & 4.88 \\
        &                         & SPECTRA & 4.42 & 14.00 & 23.53 & 4.88 \\
        \cmidrule(lr){2-7}
        & \multirow{2}{*}{Llama2} & Base    & 3.54 & 10.00 & 17.39 & 9.76 \\
        &                         & SPECTRA & 5.31 & 6.12  & 20.59 & 0.00 \\
        \midrule
        \multirow{2}{*}{\textbf{Law}}
        & \multirow{2}{*}{Llama2} & Base    & 7.96 & 18.00 & 20.29 & 7.32 \\
        &                         & SPECTRA & 5.31 & 6.00  & 14.49 & 7.32 \\
        \bottomrule
    \end{tabular}}
    \caption{Model performance on MultiChallenge \citep{sirdeshmukh2025multichallengerealisticmultiturnconversation} across domains.}
    \label{tab:spectra_comparison}
\end{table}

%% file: sections/intro/limit.tex
\section{Limitations}
The adversarial landscape surrounding language model safety is rapidly evolving, and it is likely that sufficiently adaptive attacks could be formulated to circumvent the guardrails \text{SPECTRA} provides. Generalizing the method to maintain robustness against emerging attack strategies remains an important direction for future work. Additionally, while our evaluations demonstrate that \text{SPECTRA} has negligible impact on general coherence and domain-specific knowledge, modern language models possess a wide range of capabilities that were not assessed in this work. It is possible that subtle, task-specific capabilities could be affected by weight-level steering in ways not captured by our benchmarks, and a more comprehensive capability evaluation would be necessary to fully characterize the method's impact and support broader generalization.

%% file: sections/intro/acknowledgement.tex
\section{Acknowledgment}
This material is based upon work supported by the U.S. National Science Foundation under Grant Nos. IIS-2202699, IIS-2416895, IIS-2301599, CMMI-2301601, DMS-2529302, and OIA-2616279.

%% file: sections/intro/discussion.tex
\section{Discussion}\label{sec:dis}
In this work, we introduced \text{SPECTRA}, a training-free method for restoring safety behaviors in fine-tuned language models that suffer from alignment degradation, even when fine-tuned on entirely benign data. Our evaluations demonstrate that \text{SPECTRA} substantially improves refusal behavior and robustness against a diverse set of jailbreak attacks across multiple models and domains, with no significant degradation in general coherence, domain-specific knowledge, or false refusal rates. Notably, \text{SPECTRA} also improves domain-specific safety without requiring any domain-specific safety data.
The effectiveness of \text{SPECTRA} stems from a key insight: while full-rank weight perturbation along the refusal direction is effective, it is indiscriminate and disrupts unrelated model capabilities. By restricting the intervention to an activation-aware low-rank subspace in which safety-relevant components are concentrated \citep{wei2024assessingbrittlenesssafetyalignment}, \text{SPECTRA} achieves targeted refusal steering with minimal collateral effects. Practically, this requires modifying at most two projection matrices per layer, making the method both simple and computationally lightweight.

\noindent\textbf{Future Work.} Several promising directions emerge from this work. Beyond its application to fine-tuning safety restoration, \text{SPECTRA}'s weight-level steering framework could generalize to a broader class of behavioral steering tasks, offering a training-free mechanism for eliciting or suppressing model behaviors beyond refusal. The simplicity and low computational cost of \text{SPECTRA} also make it well-suited for deployment in production systems, where it could be evaluated against a wider range of adversarial attacks and assessed across a broader set of model capabilities and application domains.

\noindent\textbf{Ethical Considerations} We believe mechanistic interpretability techniques
can alleviate many AI safety concerns and assist
in creating safe and reliable AI systems. However, dual-use remains a concern, as research in mechanistic interpretability can aid malicious intentions for exploiting/creating unsafe AI, potentially leading our methodology to be used for malicious intent. We beleive future work in this direction would aid mitigate such downsides.

%% file: sections/intro/related.tex
\section{Related Work}
\paragraph{Improving Safety of Fine-Tuned Models:} Works have aimed at improving the safety of fine-tuned language models via further preference optimization \citep{han2024medsafetybenchevaluatingimprovingmedical}, curating specific safety data to improve overall safety evaluation by further tuning \citep{jan2024multitaskmayhemunveilingmitigating}, introducing a novel finetuning objective to mitigate fine-tuning-based safety attacks \citep{qi2024safetyalignmentjusttokens}, creating safety prefix prompts as backdoors during the fine-tuning procedure \citep{NEURIPS2024_094324f3} or via separating states during the fine-tuning procedure and optimization state drift to prevent alignment degradation \citep{huang2024lazy}. All of the aforementioned methods rely on further gradient calculations/tuning or altering the fine-tuning objective or require further domain-specific calibration data , which can pose constraints on the efficacy and cost of the fine-tuning procedure. Our method distinguishes itself by directly injecting a steering vector into a low-rank safety-related space of the fine-tuned variant, making it simple, efficient, and having low spillover onto other capabilities. Erstwhile, \cite{chhabra2024neuroplasticity,jain2023mechanistically} focus on mechanistically intepreting the effects of fine-tuning on know language model circuits. 

\paragraph{Refusal Steering Vectors:} Prior work has identified the refusal steering vector \citep{arditi2023refusal} and utilized its significance to create a white box jailbreak method that incurs low cost and enjoys a high Attack Success Rate. Other works have discovered such steering vectors via the use of sparse autoencoders \citep{o2024steering}. Prior research has also utilized steering vectors to mitigate false refusals in language models \citep{wang2024surgical}. While some works have applied steering vectors to steer refusal and improve model safety \citep{obrien2025steeringlanguagemodelrefusal}, they have shown to have significant downsides on the general coherence of the language model and/or require inference time interventions \citep{lee2025programmingrefusalconditionalactivation} to steer model activations. \citet{chhabra2025understandingimprovingrefusalcompressed} also utilized steering vectors to improve the safety of compressed models.

\section{Compute Statement}
For running experiments, we utilized 6 NVIDIA A6000 GPUs with total computations for the experiments covered to be around 300 GPU hours. 

%% file: sections/appendix/alpha.tex
\section{$\alpha$ values of the reported experiments}
\label{app:alpha}
The value of $\alpha$ is model-dependent and was selected via mild hyperparameter tuning for each model variant. Table~\ref{tab:alpha} reports the values used in all experiments.

\begin{table}[h]
    \centering

    \resizebox{\columnwidth}{!}{
    \begin{tabular}{llll}
        \toprule
        \textbf{Model family} & \textbf{Domain}  & \textbf{Attention$_\alpha$} & \textbf{MLP$_\alpha$} \\
        \midrule
        \textsc{Llama-2 Chat}       & Medical    & 0.07 & 0.05 \\
        
        \textsc{Llama-2 Chat}       & Law & 0.1 & 0.03 \\ 
        \textsc{Llama-2 Chat}       & Finance  & 0.1 & 0.05  \\
        \textsc{Gemma2-9b-it}   & Medical           & 0.01& 0.00 \\
        \textsc{Gemma2-9b-it}   & Finance           & 0.01&0.00 \\
        \bottomrule
    \end{tabular}
    }
    \caption{Values of the steering hyperparameter $\alpha$ used for each model variant and domain in all experiments.}
    \label{tab:alpha}
    
\end{table}

\begin{figure}[!h]
    \centering
    \includegraphics[width=0.85\linewidth]{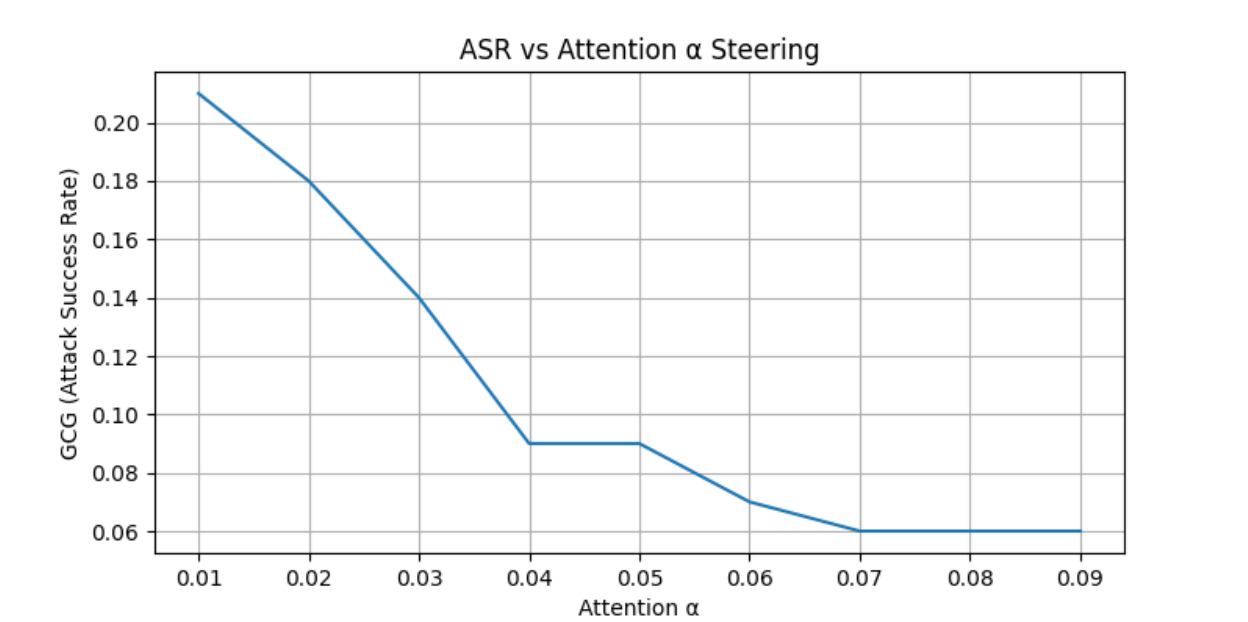}
    \caption{ASR of GCG attacks vs $\alpha$ for attention heads, with fixed $\alpha$ for MLPs.}
    \label{fig:placeholder}
\end{figure}

\begin{figure}[!h]
    \centering
\includegraphics[width=0.85\linewidth]{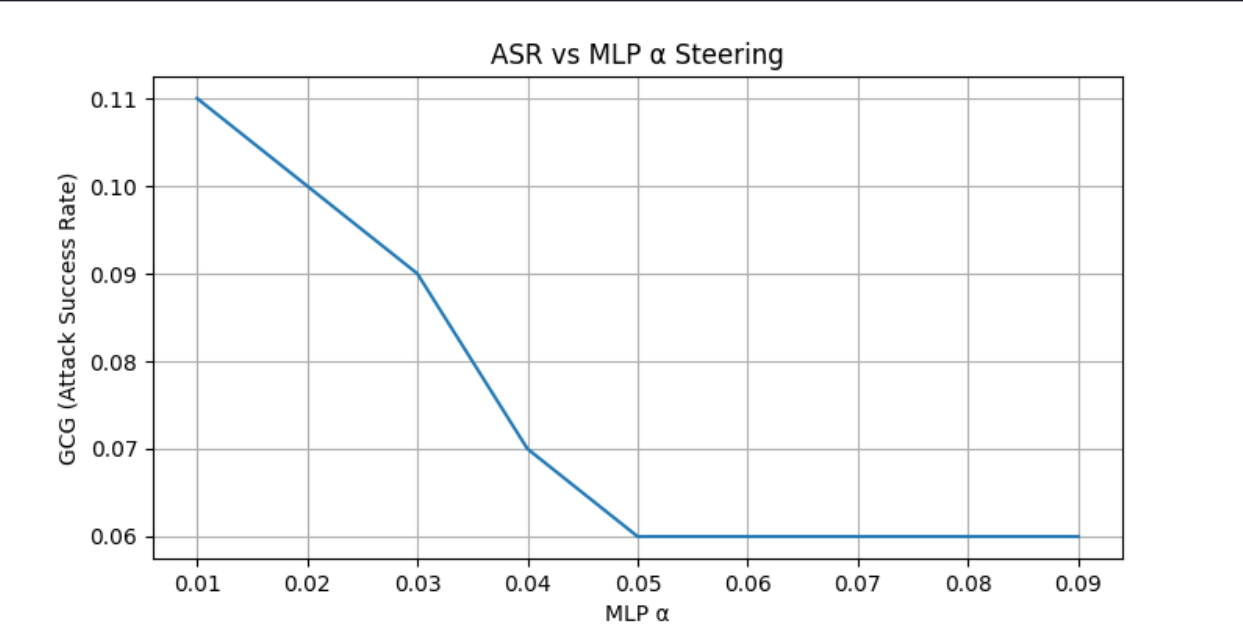}
    \caption{ASR of GCG attacks vs $\alpha$ for MLP, with fixed $\alpha$ for attention.}
    \label{fig:placeholder1}
\end{figure}

\paragraph{Hyperparameter sensitivity.} 
we perform a sweep over $\alpha$ and report GCG attack success rates \citep{zou2023universal} on the medically fine-tuned \textsc{Llama-2-Chat} \citep{rohanian2024exploring} as a function of $\alpha$. We consider two cases. In the first, we fix the MLP steering coefficient at $\alpha=0.05$ and sweep the attention steering coefficient; results are shown in Figure~\ref{fig:placeholder}. In the second, we fix the attention steering coefficient at $\alpha = 0.07$
and sweep the MLP steering coefficient; results are shown in Figure~\ref{fig:placeholder1}. 


%% file: sections/appendix/rank.tex
\section{Rank of the Low Matrix for Weight Steering}
\label{app:rank}
We report the rank $\rho$ of the low-rank projection matrices used in SPECTRA for each model variant and domain in Table~\ref{tab:rank}.

\begin{table*}[h]
    \centering
    \renewcommand{\arraystretch}{1.2}
    
    \begin{tabular}{llll}
        \toprule
        \textbf{Model family} & \textbf{Domain}  & \textbf{Attention} & \textbf{MLP} \\
        \midrule
        \textsc{Llama-2 Chat}       & Medical    & 2096/4096 & 2096/4096 \\
        
        \textsc{Llama-2 Chat}       & Law & 3096/4096 & 3096/4096\\ 
        \textsc{Llama-2 Chat}       & Finance  & 2096/4096 & 2096/4096  \\
        \textsc{Gemma2-9b-it}   & Medical           & 1048/2048 & 2048/3048 \\
        \textsc{Gemma2-9b-it}   & Finance           & 1048/2048 &2048/3048\\
        \bottomrule
    \end{tabular}
    
    \caption{Low-rank projection ranks in SPECTRA, reported as selected rank $\rho$/full rank.}
    \label{tab:rank}
\end{table*}

%% file: sections/appendix/rd.tex
\section{Refusal Direction Selecting Algorithm}
\label{sec:rdfsa}
The construction in Section 2.1 yields a family of candidate directions indexed by layer $l\in[L]$ and token position $i\in I$. To select a single, causally effective refusal direction,  we evaluate each candidate through intervention experiments on a held-out validation set. A direction is considered effective if (i) ablating it from activations induced by harmful prompts in $\mathcal{D}_{\text{harmful}}^{\text{(val)}}$ should suppress refusal, and (ii) injecting it into activations induced by harmless prompts in $\mathcal{D}_{\text{harmless}}^{\text{(val)}}$ should elicit refusal. We select the candidate $(i^*,l^*)$ that maximizes this combined causal effect, and denote the resulting direction as  $\mathbf{r}\coloneqq \mathbf{r}_{i^*}^{(l^*)}$, with normalized version $\hat{\mathbf{r}}$.
We borrow the refusal direction selection algorithm from \citet{arditi2024refusallanguagemodelsmediated}.
Given a collection of difference-in-means vectors, denoted as \( \{ \mathbf{r}_i^{(l)} | i \in I, l \in [L] \} \), we evaluate the following key metrics:

\begin{itemize}[leftmargin=*]
    \item \textbf{\texttt{bypass\_score}}: Measures the average refusal rate on the validation set of harmful prompts (\(\mathcal{D}_{\text{harmful}}^{\text{(val)}}\)) when applying directional ablation to \( \mathbf{r}_i^{(l)} \).  
    \item \textbf{\texttt{induce\_score}}: Assesses the average refusal rate on the validation set of harmless prompts (\(\mathcal{D}_{\text{harmless}}^{\text{(val)}}\)) when the activation addition of \( \mathbf{r}_i^{(l)} \) is applied.  
    \item \textbf{\texttt{kl\_score}}: Computes the average Kullback-Leibler (KL) divergence between the model's probability distributions at the final token position when evaluated on \( \mathcal{D}_{\text{harmless}}^{\text{(val)}} \) with and without directional ablation of \( \mathbf{r}_i^{(l)} \).  
\end{itemize}

To identify the optimal direction \( \mathbf{r}_{i^*}^{(l^*)} \), we select the vector with the lowest \texttt{bypass\_score}, while ensuring the following constraints are met:  

\begin{itemize}[leftmargin=*]
    \item \(\texttt{induce\_score} > 0\)  
    \begin{itemize}
        \item Ensures that the selected direction is capable of inducing a refusal response.  
    \end{itemize}
    
    \item \(\texttt{kl\_score} < 0.1\)  
    \begin{itemize}
        \item Prevents the selection of directions that excessively alter model behavior on benign prompts.  
    \end{itemize}
    
    \item \( l < 0.8L \)  
    \begin{itemize}
        \item Restricts the selection to earlier layers, avoiding interference with unembedding representations.  
    \end{itemize}
\end{itemize}

%% file: sections/appendix/GCG.tex
\section{GCG Attack Details}
We borrow and modify the methodology of \citet{wei2024assessingbrittlenesssafetyalignment} to generate adversarial suffixes, which is:
Run the GCG attack \cite{zou2023universal} for $500$ iterations,
with adversarial string initiated as ``\texttt{!!!!!!!!!!!!!!!!!!!!}" and a batch size of $256$, top-$k$ as $128$, with optimization over the fine-tuned models(including the SPECTRA Variants), with the system prompts removed, for three independent trials. We then identify the top three suffixes with the highest attack success rates on AdvBench, and use them in our evaluation.

%% file: sections/appendix/zero.tex
\section{Details of Zero-Shot Evaluations}
\begin{enumerate}[leftmargin=*]
    \item \textbf{ARC-Challenge:}
    \begin{enumerate}
        \item \textbf{Downstream Task:} Science Question Answering.
        \item \textbf{Overview:} This metric gauges model performance on the ARC-Challenge portion of the AI2 Reasoning Challenge dataset. It comprises grade-school science questions that necessitate complex reasoning and an in-depth understanding of scientific principles\footnote{Further details can be found at \url{https://allenai.org/data/arc}.}.
    \end{enumerate}
    \item \textbf{HellaSWAG:}
    \begin{enumerate}
        \item \textbf{Downstream Task:} Commonsense Reasoning.
        \item \textbf{Overview:} HellaSWAG is designed to test commonsense reasoning capabilities. It presents a context followed by several multiple-choice endings, with the objective of selecting the most plausible continuation. The dataset challenges models to interpret and reason about everyday situations\footnote{Additional information is available at \url{https://huggingface.co/datasets/Rowan/hellaswag}.}.
    \end{enumerate}
    \item \textbf{WinoGrande:}
    \begin{enumerate}
        \item \textbf{Downstream Task:} Commonsense Reasoning.
        \item \textbf{Overview:}  WinoGrande is a large-scale dataset for assessing commonsense reasoning. Presented as a fill-in-the-blank task with binary choices, the aim is to select the appropriate option, demanding robust commonsense understanding while mitigating dataset-specific biases\footnote{Further information is available at \url{https://huggingface.co/datasets/winogrande}.}.
    \end{enumerate}
    \item \textbf{BoolQ:}
    \begin{enumerate}
        \item \textbf{Downstream Task:} Yes/No Question Answering.
        \item \textbf{Overview:} BoolQ is a dataset focused on yes/no questions, featuring 15,942 naturally occurring examples. Each instance comprises a question, a passage, and the corresponding answer, with optional contextual information such as the page title. The setup is akin to text-pair classification tasks found in natural language inference research\footnote{More details are in \url{https://github.com/google-research-datasets/boolean-questions}.}.
    \end{enumerate}
    \item \textbf{RTE (Recognizing Textual Entailment):}
    \begin{enumerate}
        \item \textbf{Downstream Task:} Textual Entailment.
        \item \textbf{Overview:} The RTE task involves deciding whether a hypothesis can be logically inferred from a given premise. The dataset consists of sentence pairs, where the goal is to classify each pair as either "entailment" (if the hypothesis logically follows from the premise) or "not entailment" (if it does not)\footnote{Additional details are available at \url{https://huggingface.co/datasets/nyu-mll/glue\#rte}.}.
    \end{enumerate}
\end{enumerate}

%% file: sections/appendix/direction.tex
\subsection{Refusal Directional Change After Fine-Tuning}
\label{app:rdir}
Table~\ref{tab:rdir} reports the layer and token position of the refusal direction in both the base and fine-tuned models, showing that domain fine-tuning shifts the source position of the refusal direction across all evaluated variants.

\begin{table}[h!]
\centering
\setlength{\tabcolsep}{6pt}
\renewcommand{\arraystretch}{1.2}
\begin{tabular}{@{}llll@{}}
\toprule
\textbf{Model} & \textbf{Domain} & \textbf{$l_2/l_1$} & \textbf{$tp_2/tp_1$} \\
\midrule
Llama2  & Medical & $13/14$ & $-5/-1$ \\
        & Law     & $12/14$ & $-2/-1$ \\
        & Finance & $12/14$ & $-5/-1$ \\
\midrule
Gemma2  & Medical &  $21/31$& $-1/-1$  \\
        & Finance & $31/31$ &  $-5/-1$\\
\bottomrule
\end{tabular}
\caption{Layer and token position of the refusal direction in the base model ($l_1$, $tp_1$) and fine-tuned model ($l_2$, $tp_2$) for each variant.}
\label{tab:rdir}
\end{table}

%% file: sections/appendix/ref_data.tex
\section{Datasets for Refusal Direction Estimation}
\label{sec:ref_data}
Following \citet{arditi2024refusallanguagemodelsmediated}, we form $(\mathcal{D}_{\text{harmful}},\mathcal{D}_{\text{harmless}})$ for estimating refusal directions. We construct $\mathcal{D}_{\text{harmful}}$ from harmful instructions in \textsc{AdvBench} \citep{zou2023representation}, \textsc{MaliciousInstruct} \citep{huang2023catastrophic}, \textsc{TDC2023} \citep{mazeika2024harmbench}, and \textsc{HarmBench} \citep{mazeika2024harmbench}, and construct $\mathcal{D}_{\text{harmless}}$ from \textsc{Alpaca} \citep{taori2023alpaca}. Each set contains 160 samples, split into 128 training and 32 validation examples, which we use to compute candidates and select the final normalized refusal direction $\hat{\mathbf r}$.